\documentclass[letterpaper, 10 pt, conference]{ieeeconf}  

\usepackage{amssymb}
\usepackage{graphicx,amsmath}
\usepackage{lineno}
\usepackage{cite}
\usepackage{caption,url}
\usepackage{hyperref}
\usepackage{float}
\usepackage{epstopdf}
\usepackage{authblk}
\usepackage{dblfloatfix}
\usepackage{multirow}
\usepackage{algpseudocode}
\usepackage{algorithm}
\usepackage{xcolor}

\IEEEoverridecommandlockouts                              

\title{\LARGE \bf Learning Multimodal Bipedal Locomotion and \\ Implicit Transitions: A Versatile Policy Approach }

\author{Lokesh Krishna and Quan Nguyen
\thanks{Lokesh Krishna and Quan Nguyen are with the Department of Aerospace and Mechanical Engineering, University of Southern California,
        Los Angeles, CA 90089, USA
        {\tt\small lkrajan@usc.edu, quann@usc.edu }}%
}

\begin{document}

\maketitle
\thispagestyle{empty}
\pagestyle{empty}

\begin{abstract}

In this paper, we propose a novel framework for synthesizing a single multimodal control policy capable of generating diverse behaviors (or modes) and emergent inherent transition maneuvers for bipedal locomotion. In our method, we first learn efficient latent encodings for each behavior by training an autoencoder from a dataset of rough reference motions. These latent encodings are used as commands to train a multimodal policy through an adaptive sampling of modes and transitions to ensure consistent performance across different behaviors. We validate the policy's performance in simulation for various distinct locomotion modes such as walking, leaping, jumping on a block, standing idle, and all possible combinations of inter-mode transitions. Finally, we integrate a task-based planner to rapidly generate open-loop mode plans for the trained multimodal policy to solve high-level tasks like reaching a goal position on a challenging terrain. Complex parkour-like motions by smoothly combining the discrete locomotion modes were generated in  $\sim$$3$ min. to traverse tracks with a gap of width $0.45$ m, a plateau of height $0.2$ m, and a block of height $0.4$ m, which are all  significant compared to the dimensions of our mini-biped platform.


\end{abstract}


\section{Introduction}
\label{sec:intro}
Model-Free Reinforcement Learning (RL) has emerged as an effective alternative to classical and optimization-based techniques for controller synthesis in the paradigm of legged robots \cite{ siekmann21, gabe22rapid, Lee2020}.
By theory, RL as an optimization technique does not suffer from any modeling constraints and should be general by design in learning multiple behaviors. Nonetheless, due to practical limitations like reward design, task specification \cite{pmlr-v164-agrawal22a}, and catastrophic forgetting \cite{crcf2021}, RL for robot locomotion has been shown to generate policies that specialize in a single behavior such as walking \cite{gabe22rapid, Lee2020} or a conservative set of periodic motions \cite{siekmann20}. 
In contrast, a model-based control pipeline akin to that of Boston Dynamics' Atlas \cite{bd2023} has been shown to generalize to diverse behaviors in well-choreographed routines like parkour, dance, and loco-manipulation in controlled environments. 
However, a direct counterpart is absent in the paradigm of RL-based control for bipedal locomotion, which is the focus of this work.

An ideal RL policy for multimodal locomotion is expected to learn multiple behaviors/modes and transition maneuvers between those behaviors. Learning transitions raises a two-fold challenge: 1) lack of explicit transition demonstrations, unlike for different behaviors, and 2) the number of transitions increasing as the square of the number of behaviors. The former limitation requires the policy to develop emergent transition maneuvers,  while the latter makes it harder to guarantee consistent performance across all possible transitions. Thus, we require the policy to learn the different behaviors explicitly (with reference demonstration) and the inter-behavior transitions implicitly (without reference demonstration) while ensuring consistent performance across the $n$ behaviors and the $n^2$ transitions. 

For behavior generalization in RL, an alternative would be to learn multiple policies, where each specializes in a single behavior, and later switch between them to solve high-level tasks. Apart from the apparent computational limitation of scaling to an increased number of behaviors, designing smooth and feasible transition strategies between these different policies trained to operate in distinct and possibly disjoint local regions of the state space is non-trivial. On the other hand,  techniques from  Multi-Task RL (MTRL) \cite{impala18} have been developed to generalize across diverse environments and cognitive tasks. 
Unlike unstructured cognitive tasks, for learning versatile locomotion, we can exploit the specific fixed structure for the problem. That is, irrespective of the behavior we desire, the environment (a hybrid dynamical system with contacts), observation (state of the dynamical system), and action(torques/control) spaces remain the same.
Thus the synthesis of multimodal locomotion only requires us to identify a general training strategy. In agreement with our assertion, \cite{Li2022} demonstrates the extraction of kinematic primitives from policies trained on a single behavior, thereby realizing novel behaviors through human-AI shared autonomy. This work further motivates us to explore the possibility of realizing multiple behaviors through a single policy while learning feasible inter-behavior transitions. 
\begin{figure}[t]
	\centering
         \includegraphics[width=\linewidth]{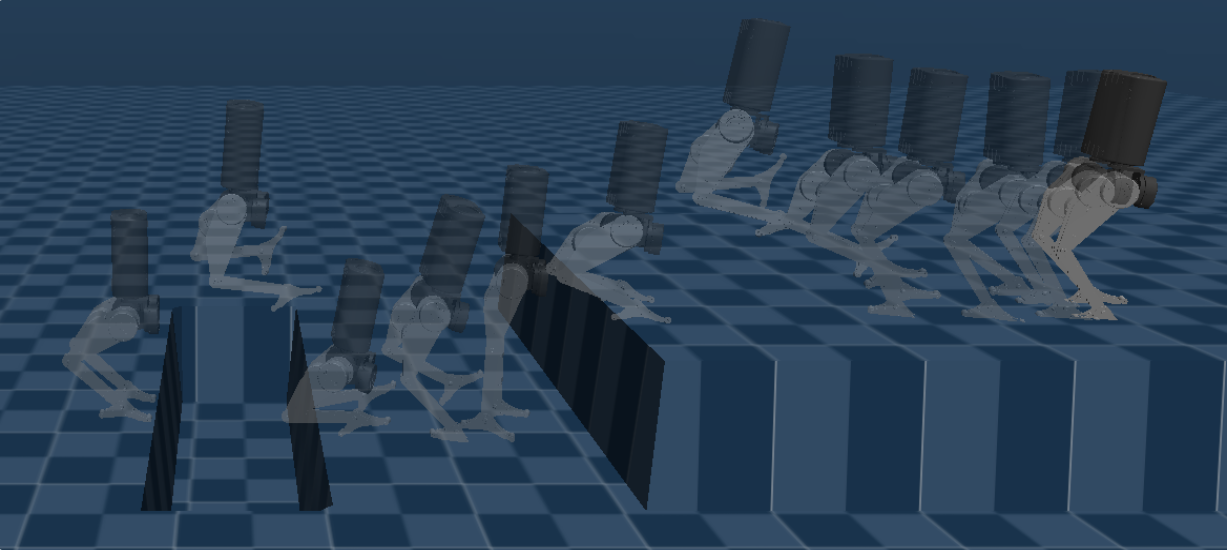}       
        
	\caption{\textit{A parkour-like  behavior composed in $\sim 3$ mins by the proposed approach to traverse a track with a gap of width $0.45$ m and block of height $0.4$ m. Accompanying video results can be found at \href{https://youtu.be/-E5EBf8695A}{\color{blue}https://youtu.be/-E5EBf8695A}}}
      \label{fig:diff_mode_motion_trace}
	 \vspace{-7mm}
\end{figure}

A major challenge for RL in realizing multi-behavior policies is catastrophic forgetting and remembering \cite{crcf2021}, which hinders the consistent performance of a multi-skilled policy, causing mode collapse and aliasing, respectively. \cite{Vollenweider2022} tackles it by learning all the skills in parallel,  whereas \cite{Peng2022} proposes to learn a sequence of repeated skills. However, neither of the above considers the inherent bias in the complexity of different skills. For instance, a skill like a backflip may be more complicated than flat-ground walking, thus requiring more samples to master. Additionally, equal exposure to skills of varied complexity results in a local optimum where the policy is successful only in a subset of the skills, causing mode collapse. \cite{body_as19} tackles a similar variance in complexity across body shape variations through an adaptive sampling of the body shape parameters to ensure the more frequent encountering of challenging variants in training, thereby leading to a consistent policy.

 In humanoid character control, \cite{football2021, Peng2022} show impressive locomotion skills learned from human motion capture data, but their dependence on the fidelity of the demonstrations is unclear. The target system (a humanoid character) and the demonstration expert (a human) are morphologically alike, with the same form factor and similar dynamics. However, such a rich collection of dynamically-consistent motion capture data is unavailable for robots of varied form factors and morphologies. A common solution is to generate practical demonstrations from a model-based framework, such as Trajectory Optimization (TO), for RL policies to imitate \cite{bogdanovic22}, which is redundant as two full-order optimizations are performed concurrently for the same objective.
Additionally, using overspecified rewards and differences in modeling choices can limit RL by TO's performance. Typically, TO generates dynamically-plausible trajectories based on a rough initial guess, such as linear interpolation and waypoints \cite{choung21}, which are then realized by Model Predictive Control (MPC) based controllers. In contrast, since RL directly synthesizes control policies, we ponder the possibility of learning a policy to realize locomotion modes from a library of rough motion demonstrations. Recently \cite{Li2022rough} showcased agile skills through adversarial imitation of rough partial demonstrations, providing a critical insight that the physical plausibility of the reference trajectories is not vital for the success of RL.

For quadrupedal locomotion, \cite{gabe22walk} demonstrated multiple variants of a single behavior (walking) by inputting an explicit behavior specification as an input to the policy, which is tuned during inference to adapt to unseen terrain conditions. Having been designed with periodic gait parameters, extending the above behavior specification to represent an aperiodic transient motion like jumping on a block is unclear. Alternatively, \cite{Li2022} uses one-hot encoding to represent and realize multiple locomotion skills through a  single control policy capable of transient behaviors. Such a naive choice of behavior encoding strategy 1) increases the dimension of the command vector for every new behavior added and 2) bears no fruitful information regarding the properties of the commanded motion, which can be vital while composing a smooth composition of multiple behaviors, as in parkour. To this end, we propose a technique to automatically generate latent encodings (called latent modes)  with a fixed dimension using autoencoders (called mode encoders) purely based on a demonstration dataset. 
Similar to template planning \cite{De2022}, using these latent modes, we propose mode planning to rapidly realize complex behavior compositions instead of performing full-order TO.  


In the proposed direction, \cite{Li2022}  demonstrates a handful of behaviors ($3$ in their case) and discreetly switched transitions on a wheeled quadrupedal robot.
However, behaviors with significant flight phases are absent, which are time critical and hence call for precise transitions. For example, a transition from hop to walk \cite{siekmann20} may be trivial, but in a parkour-like motion where you run, jump and land on a block of comparable body height, the switching time and strategy becomes highly crucial to ensure a successful landing, which we also investigate in this work. Thus the primary contributions of our paper are as follows.

\begin{itemize}
    \item A general framework to synthesize a single versatile multimodal policy capable of performing diverse locomotion modes, broadly classified as
    \begin{itemize}
        \item Periodic Modes: walk, hop, leap, etc
        \item Transient Modes: launch and land on a block
        \item Steady-state Modes: stay idle in nominal rest pose
    \end{itemize}
    while learning feasible inter-mode transition maneuvers implicitly.
    \item A novel adaptive sampling technique to ensure consistent performance across locomotion modes and transitions by addressing mode collapse.
    \item A mode planner for rapidly generating complex behavior compositions to enable the trained multimodal policy to solve fruitful high-level tasks.
\end{itemize}

\begin{figure*}[t!]
	\centering

	\includegraphics[width=170mm]{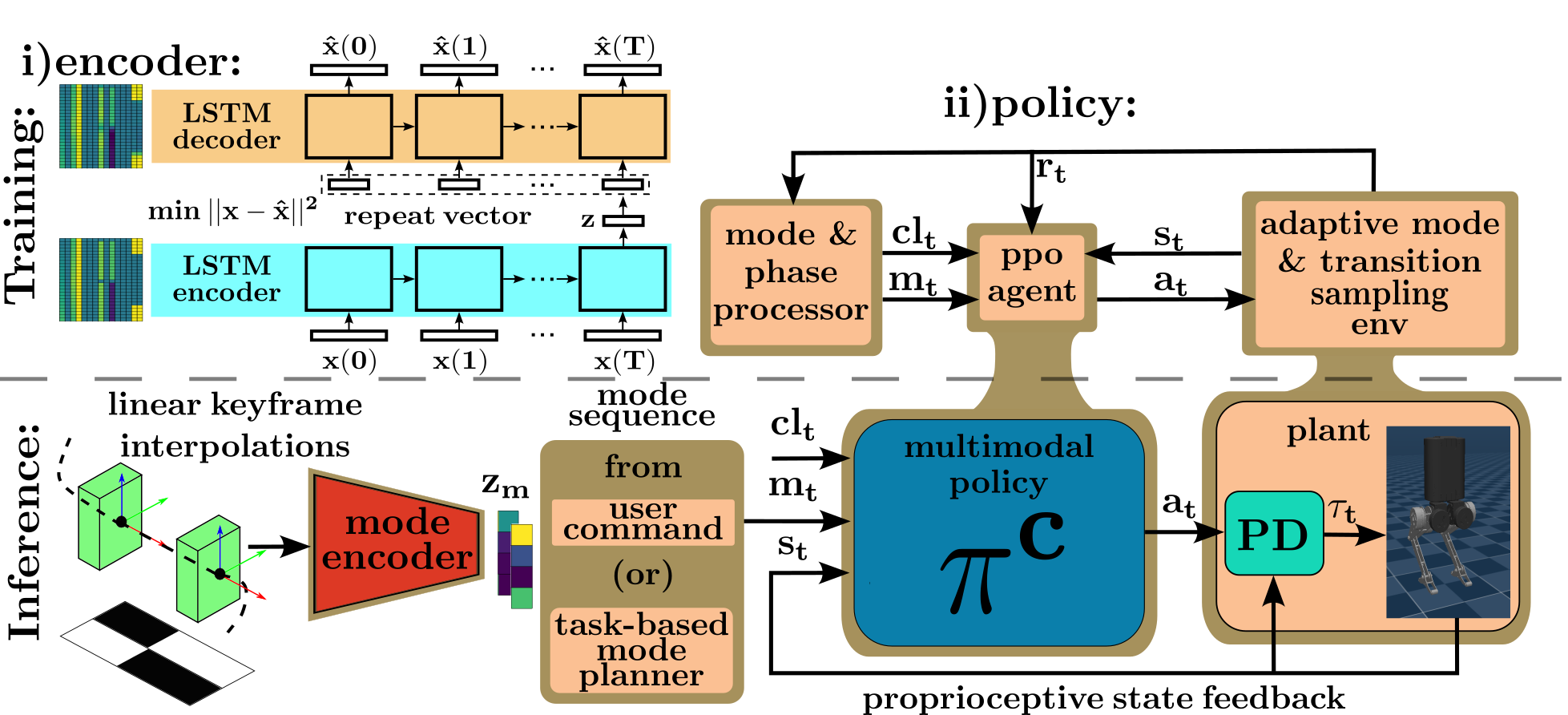}	
	\caption{\textit{The decoupled training procedure for i) encoder and ii) policy  (top) and the control pipeline for inference (bottom) }}
	\label{fig:control_arch}
	\vspace{-5mm}
\end{figure*}

\section{Proposed Approach}
\label{sec:method}


This section describes our proposed approach for learning multiple locomotion modes and inter-mode transitions in a single policy, followed by composing mode plans to solve high-level tasks. Fig \ref{fig:control_arch} shows a pictorial representation of the framework with the proposed training and inference pipelines. Our framework consists of three modules:
\begin{itemize}
    \item a mode encoder: to map reference motions to a low-dimensional latent space.
    \item a multimodal policy: to learn optimal control strategies for distinct locomotion modes.
    \item a mode planner: to generate optimal mode plans and solve high-level tasks. 
\end{itemize}
First, we train the mode encoder to learn efficient encoding (latent modes) of a repertoire of reference motions by minimizing a reconstruction loss. Using these latent modes, we train a multimodal policy to imitate the commanded locomotion modes sampled adaptively to prevent 1) failure in a subset of modes: mode collapse and 2) overlapping of distinct modes: mode aliasing. With the trained multimodal policy, we use a mode planner to solve high-level tasks through open-loop mode planning. Each of the above modules is explained in detail as follows.

\subsection{Generating Rough Reference Motions}
\label{ssec:denonstration}

\begin{figure}
	\centering
   	\includegraphics[width=\linewidth,height=6cm]{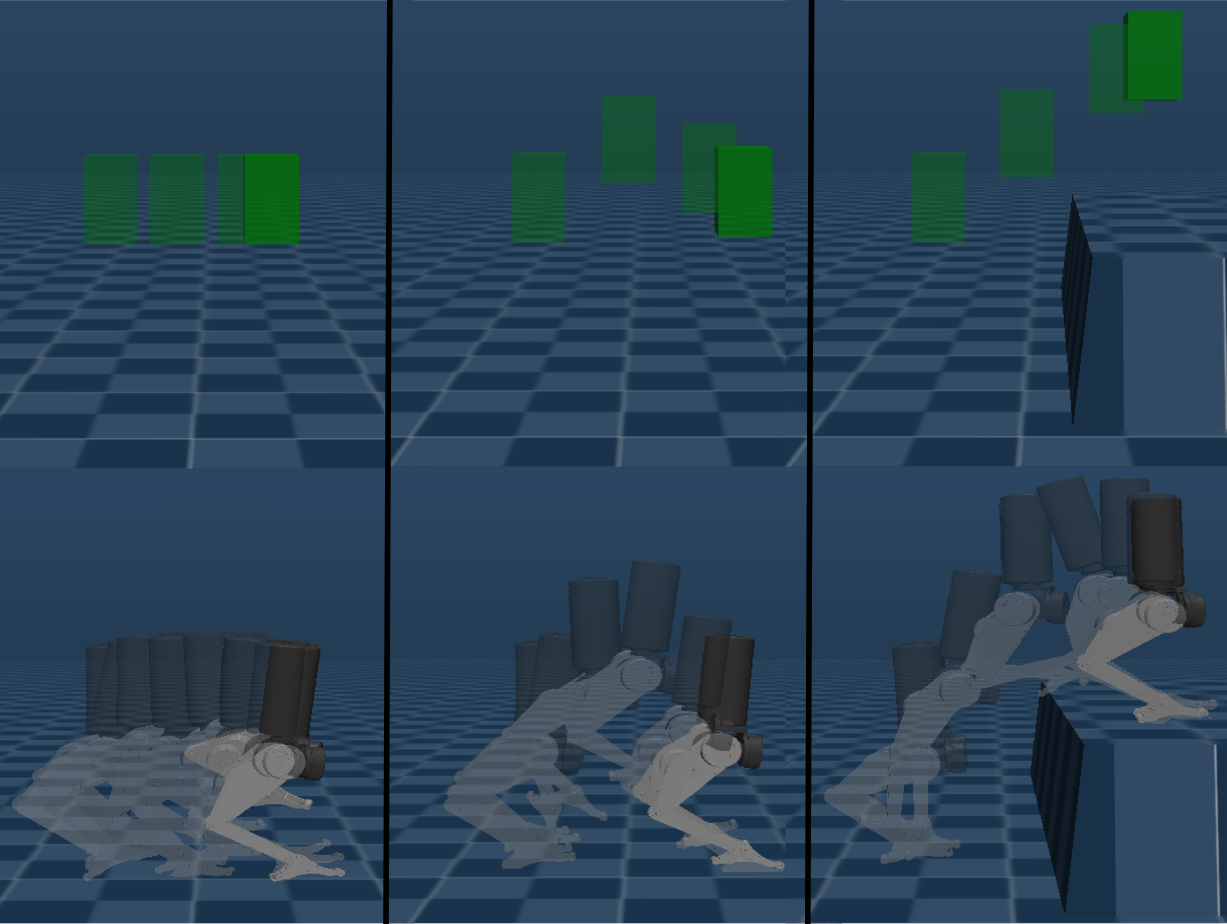}	
	\caption{\textit{The locomotion modes (from left to right): walk, leap, and launch with the rough keyframe references (top) and corresponding actual motion by the policy (bottom).}}
	\label{fig:mt_modes}
	\vspace{-6mm}
\end{figure}

 To preserve the scalability of our approach to diverse behaviors while also being agnostic to the robot's morphology, we use rough reference motions, which are not required to be either kinematically or dynamically feasible but merely be a visual proxy of the desired motion. We only need the keyframes of the base pose (position and orientation) and, optionally, the contact sequence (for well-defined contact patterns) in a desired motion's reference. As shown in Fig \ref{fig:mt_modes} (top), we generate the rough reference motion by linearly interpolating between user-defined keyframes. 
We can optionally specify the reference contact sequence through a binary signal for each leg to obtain locomotion modes with well-defined contact patterns. For instance, a reference for walking mode defined as constant base height with increasing position value along the heading direction seldom results in a distinct left-right marching motion, making it hard to distinguish between walking and hopping. Note that there are no reference trajectories for the actuated states (joint DOFs), thus allowing the policy to learn the required actions to realize without overspecified constraints.


\subsection{Mode Encoder}
\label{ssec:mode_encoder}

Unlike prior works, \cite{pmlr-v119-hasenclever20a,football2021, Peng2022} which have a one-to-one correspondence between the agent's state and latent trajectories, we propose to encode a complete reference motion into a single point in the latent space before the agent's training. By decoupling the temporal correspondence, the policy has direct access to the compressed past and future intent, which is agnostic to its current state. As shown later in the paper, a state-agnostic mode command helps us generate complex behaviors (as in Fig. \ref{fig:diff_mode_motion_trace}) with discrete switching through simple mode planning methods. 
Since the reference motions are time-series data with a strong temporal relationship between each point in the trajectory, we use a Long Short-Term Memory
(LSTM) Autoencoder to learn our mode embeddings. Additionally,  the architecture of an LSTM network is independent of the input trajectory's length, which is of practical significance. Unlike periodic (walk) or steady state (stay idle) modes, where we can generate reference motions of arbitrary length, transient motions such as jumping on a block are time-critical, thus having a  defined trajectory length. As shown in Fig \ref{fig:control_arch} (top) , we train an LSTM Encoder, $E_\theta$ and Decoder, $D_\psi$ pair based on the reconstruction loss below  

\begin{equation}
    \underset{\theta, \psi }{\text{min}} \|X - \hat{X}\|_2
    \vspace{-2mm}
\end{equation}

\begin{figure*}[t]
	\centering
        \includegraphics[width=174mm,height=4cm]{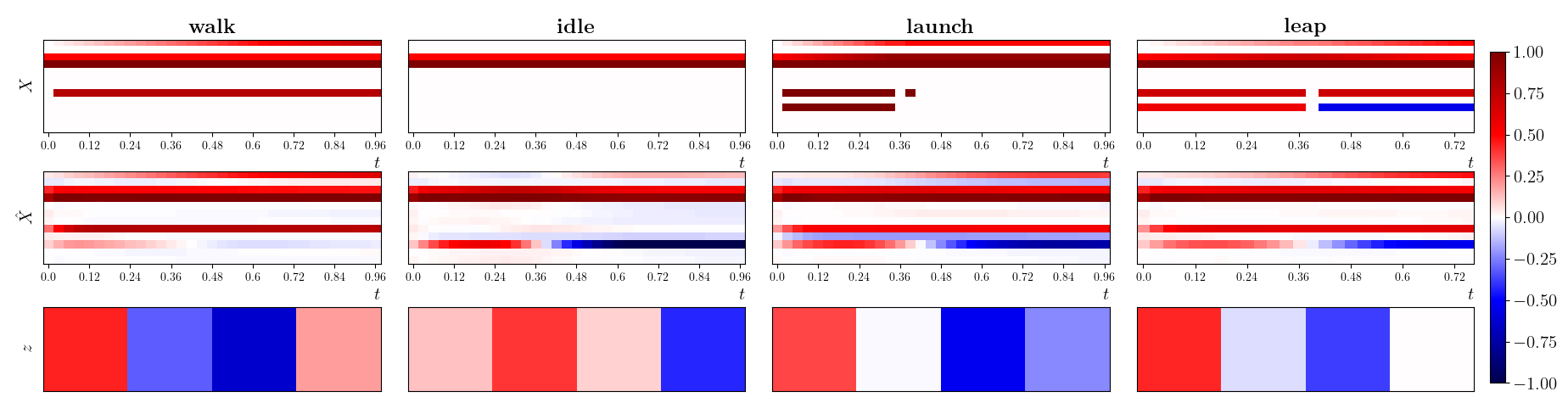}
	\caption{\textit{The rough reference motions (top) with the corresponding reconstructions (centre) and learnt embeddings(bottom) }}
	\label{fig:encoding}
	\vspace{-5mm}
\end{figure*}

where $X$ is a reference motion of length T and $\hat{X}$ is its reconstruction. Thus, $E_\theta$ recurrently takes in the entire trajectory one timestep at a time and generates a single latent vector $z$ as output. A repeat vector is then constructed, making T identical copies of $z$ and sent as an input to the decoder along with the hidden state being propagated internally in a recurrent fashion, thus outputting a reconstruction of the $X$ trajectory, $\hat{X}$ of the same length. We use the open-source implementation sequitur\footnote{ \href{https://github.com/shobrook/sequitur}{https://github.com/shobrook/sequitur}} for training the autoencoder, with identical encoder and decoder networks with a single hidden layer of 32 neurons and no activations. As seen Fig. \ref{fig:encoding} the mode encoder effectively encodes a complete trajectory into a single latent vector thereby leading to a compact representation of the desired behaviors.

\subsection{Multimodal Policy}
In contrast to traditional locomotion policies for legged robots that are command-conditioned on the base's lateral, longitudinal, and yaw velocities, we use a latent-conditioned policy, $\pi(a|s,z)$ where $z \in \mathbb{Z}$, is a latent space of behaviors/modes. Prior approaches simultaneously discover and learn such skills from scratch \cite{sharma19} or morphologically accurate motion capture data \cite{Peng2022, npmp18, football2021}, which are unsuitable for robot control. Thus we decoupled the training first to learn the latent mode encodings (as explained in Sec. \ref{ssec:mode_encoder}, \ref{ssec:denonstration}) and then trained a policy conditioned to those encodings to imitate the corresponding locomotion mode. We control the policy only through mode commands and relax the need to accurately track velocity commands as it is often merely a surrogate objective used to satisfy high-level task objectives. An indirect way to modulate the robot velocity and a method to plan compositional locomotion tasks like parkour is discussed in Sec \ref{sec:results}. Described below are the specific details pertaining to the policy.

\subsubsection{Observation Space} The observation space to our policy, $o_t$ consists of a subset of the robot's proprioceptive state feedback $s_t$, a clock signal $cl_t$, and the mode command (as a latent mode) $m_t$. Since any locomotion mode is invariant to the translational DOFs in the transverse plane (typically denoted by x and y), $s_t \in \mathbb{R}^{31}$ comprises the complete state of the robot dynamics exempting the free-floating torso's x and y position in the world coordinates. The orientation is expressed as quaternions, the translation velocity of the torso link is transformed to the base frame, and the torso's z is sent relative to the current support plane. As mentioned, $m_t$ is a latent vector encoding the given locomotion mode's reference motion. While  $m_t$ provides information on the desired locomotion mode, the policy has no explicit feedback on its progress along that motion. To this end, we define a phase variable, $\phi$, which runs from $0 \text{ to } 1$, marking the start and end of a given mode and construct $cl_t = [\sin{\phi},\:\cos{\phi}]$. Thus, $o_t = [cl_t,\:m_t,\:s_t] $ is a $33 + n_m$ dimensional vector where $n_m$ is length of the latent mode vector. While $n_m$ can be of arbitrary length, we empirically found $n_m =4$ to be the minimum sufficient value for our results.

\subsubsection{Action Space} Similar to the conventional practices, we define the action space, $a_t \in \mathbb{R}^{10}$ to be motor targets to the $10$ actuated DOFs ($5$ joints per leg) of the robot. The final applied motor torques, $\tau$ are then given by 
\begin{equation}
    \tau = K_p(a^j_t - q^j_t ) - K_d(\dot{q}^j_t)
\end{equation}

where $j,q^j_t$ are the index and position of a given joint, $K_p = 30 \text{ and } K_d=0.5$ are fixed joint PD gains. We clip $\tau$ at $30$ Nm, which is the torque limit of our actual robot but do not apply any joint angle limits to $a_t$, as it is not strictly the position target and can only be intuitively interpreted so. The feasibility of our motions is unaffected by the absence of joint angle limits as we do not perform position control but directly send torque commands. 

\subsubsection{Reward Function} With a well-defined objective of imitating the desired torso states and optionally a contact sequence, we use the following reward function. 

\begin{equation}
    r_t = w_p e^{-k_p \| er_p \|_2} + w_o e^{-k_o \| er_o \|_2} + w_c e^{-k_c \|er_c \|_2}
\end{equation}

where $w = [w_p, w_o, w_c]$ are the scaling gains, $k = [k_p, k_o, k_c]$ are the sensitivity gains, $er_p, er_o$ are the error in base position and orientation and $er_c$ is the error in contact state. The possible values of contact states are $[0,0],[1,0], [0,1],[1,1]$ which correspond to flight, single support(left), single support(right) and double support phases respectively.
The $w$ used for each experiment can be found in table \ref{table:trng_conf}, while the sensitivity gains are kept at $k = [5, 5, 2]$ for all experiments. 

\subsection{Adaptive Mode and Transition Sampling}

With the structure of our policy extablished, we now explain our proposed adaptive sampling-based training in Algo. \ref{alg:trng} to realize a consistent performance across modes and transitions. 
Unlike prior approaches of curriculum learning to gradually learn the increasing difficulty of a single behavior \cite{gabe22rapid}, comparing the complexities across different locomotion modes and defining an optimal learning order is not straightforward. However, there is indeed an inherent difference in complexity 
that requires more exposure to difficult instances while training. Hence, we propose to adaptively sample the mode commands in an episode such that the mode and the transition in which the policy performs the poorest are chosen more frequently. As shown in Algo. \ref{alg:trng}, during each episode, we choose an initial mode $\text{m}_\text{i} $ and a final mode $\text{m}_\text{f}$,  based on skewed distributions  $\text{p}_\text{i}$  and $\text{p}_\text{f}$ respectively.   $\text{p}_\text{i}$,  $\text{p}_\text{f}$ are constructed based on the records of the policy's returns across each mode ($\text{R}_\text{i}$) and each transition ($\text{R}_\text{f}$). Note that while $\text{p}_\text{i}$ provides the marginal probability of choosing $\text{m}_\text{i}$, $\text{p}_\text{f}$ provides the conditional probability of choosing $\text{m}_\text{f}$ given $\text{m}_\text{i}$. We compute the  switching timestep $\text{t}_\text{s}$, based on a randomly chosen clip\footnote{a single cycle of a locomotion mode}: $\text{clip}_\text{s}$  and a phase: $\phi_\text{s}$. The above sampling technique introduces two parameters, $\gamma$ and $\epsilon$, for each distribution, $\text{p}_\text{i}$ and $\text{p}_\text{f}$. $\gamma$ prevents the distributions from being myopic (only dependent on the previous episode's returns) and rather be computed based on the history of returns corresponding to a given mode/transition, preventing the policy from alternatively learning and unlearning. Unlearning or catastrophic forgetting can also arise due to a learnt mode not being sampled further down the training. To alleviate this, we use $\epsilon$ to add a small offset, thereby ensuring the mode with the best performance has a non-zero, yet the smallest probability of being chosen. For the results below, the values of  $\epsilon$ = $0.2$ and $\gamma_i,\gamma_f$ varies as $0.2, 0.4, 0.8$ across different training. 

\begin{algorithm}
\caption{Adaptive Sampling based Training}\label{alg:trng}
\begin{algorithmic}[1]
\Require $B = \{B_\text{1},...,B_\text{n}\}$
\Statex $\text{dataset of \lq n\rq linear keyframe interpolated demonstrations}$

\Statex  $\textbf{i) encoder training:}$
\State  $\text{train an LSTM Autoencoder on }B \text{ as described in Sec }$ 
\Statex $\text{\ref{ssec:mode_encoder}, to obtain latent encodings: } z_\text{m} = \{\text{m}_\text{1}, ..., \text{m}_\text{n}\}$ 
\Statex $\text{corresponding to the motions } \{B_\text{1},..., B_\text{n}\}$

\Statex  $\textbf{ii) policy training:}$
\State $\pi^c \gets \text{initialise policy}$

\State $\text{initialise initial mode returns, R}_\text{i}\text{(m}_\text{i}\text{)}=0, \text{R}_\text{i} \in \mathbb{R}^{n \times 1}$

\State $\text{initialise final mode returns, R}_\text{f}\text{(m}_\text{i}\text{,m}_\text{f}\text{)}=0, \text{R}_\text{f} \in \mathbb{Re}^{n \times n}$

\While{not done}

\State $\text{p}_\text{i}   = \text{RETURNS2PROB(R}_\text{i}\text{)}$
\State $\text{m}_\text{i}   \sim \text{p}_\text{i}\text{(m)}$
\State $\text{p}_\text{f}   =\text{RETURNS2PROB(R}_\text{f}\text{(m}_\text{i}\text{))}$

\State $\text{m}_\text{f}   \sim \text{p}_\text{f}\text{(m $\mid$ m}_\text{i}\text{)}$

\State $\phi_\text{s} \sim \text{U\{0.25, 0.5, 0.75\}}$
\State $\text{clip}_\text{s} \sim \text{U\{0, 1\}}$

\State $\text{compute t}_\text{s} \text{ based on } \phi_\text{s},\text{clip}_\text{s}$

\State $\text{m}_\text{sequence} =  \begin{cases}
                       \text{m}_\text{i} &  0 < \text{t} \leq \text{t}_\text{s} \\
                       \text{m}_\text{f} &  \text{t}_\text{s} < \text{t} \leq \text{T}  \\ 
                 \end{cases}\nonumber $

\State $\zeta \gets \{ 
                        \text{(o}_\text{t},\text{a}_\text{t},\text{r}_\text{t}\text{)}^\text{t=T-1}_\text{t=0}, \text{o}_\text{T} \text{\}, o}_\text{t} = \text{[cl}_\text{t},\text{m}_\text{t},\text{s}_\text{t}]^T $
\Statex $\text{ roll-out with } \pi^\text{c}\text{(a$\mid$s,m,cl)} \text{ and m}_\text{sequence}$

\State $\text{perform PPO update of } \pi^c \text{ using }\zeta$
\State $\text{R}_\text{i}\text{(m}_\text{i}) \gets \gamma_\text{i}\text{R}_\text{i} \text{+} \sum_\text{t=0}^\text{T-1} \text{r}_\text{t}$

\State $\text{R}_\text{f}\text{(m}_\text{i}\text{,m}_\text{f}\text{)} \gets \gamma_\text{f}\text{R}_\text{f}\text{(m}_\text{i}\text{,m}_\text{f}\text{)} \text{+} \sum_\text{t=0}^\text{T-1} \text{r}_\text{t}$

\EndWhile

\Function{returns2prob}{\text{R}}  \Comment{R is a vector} 
    \State $\text{k} \gets \text{-R}$ \Comment{all operations are element-wise}
    \State $\text{k} \gets \text{k - min(k)} $
    \State $\text{k =}  \begin{cases}
                       \dfrac{\text{k}}{\text{max(k)}} &  \text{max(k) $\neq$ 0} \\
                       \text{0} &  \text{else}  \\ 
                 \end{cases}\nonumber $
    \State $\text{k} \gets \text{k + } \epsilon$
    \State $\text{p} =  \begin{cases}
                       \dfrac{\text{k}}{\sum_\text{i=0}^\text{n} \text{k}} &  \sum_\text{i=0}^\text{n} \text{k} \neq \text{0} \\
                       \text{U\{0,1, ... n\}} &  \text{else}  \\ 
                 \end{cases}\nonumber $
    \State \Return \text{p}
\EndFunction

\end{algorithmic}

\end{algorithm}

\subsection{Task-based Mode Planner}

Having learnt a policy that can successfully realize a multitude of locomotion modes and transitions, we now showcase a method to rapidly compose mode plans to solve high-level tasks such as traversing a non-trivial terrain that calls for parkour-like locomotion. In this work, we use a rudimentary task-based mode planner as a proof of concept to show the ease of composing complex locomotion maneuvers with a trained multimodal policy. We parameterize the mode plan as a zero-order spline (as in Fig. \ref{fig:mode_plan}) and formulate a collocation problem by defining knots of the spline at discrete time intervals. Thus, the planning problem is to find the optimal sequence of modes to maximize a task objective using the multimodal policy. We consider the reaching task where the objective is for the robot to reach a fixed goal position in the world frame. To solve this optimization, we use tabular Q-learning to obtain an optimal policy: the mode plan. The task MDP has the knot index as the state and the mode index at that knot as the action, with the reward function defined as

\begin{equation}
    \underset{m_0,...,m_k }{\text{max}} \sum_{t=0}^{kdt}  e^{-\| p_x^{goal} - p_x(t) \|} 
\vspace{-1mm}
\end{equation}

where, $p_x^{goal}, p_x(t)$ are the goal and current base position, $k$ is the total number of knots and $dt$ is the discretisation period for the planner. The optimal plan is obtained through Monte Carlo rollouts with a given terrain type in the environment. Note that the mode planner has neither the exteroceptive feedback of the underlying terrain nor the proprioceptive feedback of the robot states. Thus the planner only generates an open-loop time-dependent plan of modes to be performed by the multimodal policy at the low level. A summary of the experiments conducted is tabulated in table \ref{table:qplans}, where we ran five ideal trials of 100 episodes in parallel, and the best plan was chosen for each task. 
\section{Results}
\label{sec:results}

\begin{table}
  \centering
  \begin{tabular}{|c|c|c|c|}
    \hline

    \multirow{2}{*}{policy}
    & modes in & no. modes,  &  reward \\
    & training &transitions  & gains, $w$ \\  
\hline
    \multirow{2}{*}{$\pi_1$}
    & idle, walk(f,b,l,r), & 9, 81 &  $[0.5, 0.5, 0]$ \\  
    & leap(f,b,l,r) &   &   \\  
\hline
    \multirow{2}{*}{$\pi_2$}
    & idle, walk(f only), & 4, 16 &  $[0.5, 0.5, 0]$ \\  
    & leap(f only), launch &   &   \\
\hline
    \multirow{2}{*}{$\pi_3$}
    & idle, walk(f only), & 4, 16 &  $[0.35, 0.35, 0.3]$ \\  
    & hop(f only), leap(f only) &   &   \\  
\hline

  \end{tabular}
  \caption{Training configuration for different experiments, where f, b, l, r stand for the front, back, left and right depicting the direction of motion w.r.t the robot}
\label{table:trng_conf}

\end{table}

In this section, we now present the exhaustive list of simulation results. We use the MuJoCo physics engine for simulation and a custom implementation\footnote{https://github.com/osudrl/RSS-2020-learning-memory-based-control} of PPO for training. 
Each training is run for a wall-clock time of two days when deployed with $8$ parallel CPU workers. First, we showcase the consistent performance of the multimodal policy across modes and transitions, followed by integration with a mode planner for seamlessly composing mode plans to solve high-level reaching tasks.

\subsection{Performance across diverse modes}
In order to highlight our key contributions, we train three different multimodal policies, namely $\pi_1 \text{, } \pi_2 \text{, and } \pi_3$ whose training configurations are tabulated in table \ref{table:trng_conf}. For comparison, we conduct each of the experiments with both uniform sampling ($\text{m}_\text{i,f}   \sim \text{U}_\text{i,f}$) and our proposed adaptive sampling of modes and transitions ($\text{m}_\text{i,f}   \sim \text{p}_\text{i,f}$). The attached video shows that the proposed framework successfully realizes the different locomotion modes, such as periodic: walk, leap, transient: launch, and steady-state: idle, from the rough reference demonstrations. 
Quantitatively the mean performance across modes is computed as the mean normalized returns over the rollouts for each mode, as shown in table \ref{table:mpamt}. We conduct the following comparisons to study the implications of our proposed design choices.

\begin{figure}[t]
	\centering

	\includegraphics[width=\linewidth]{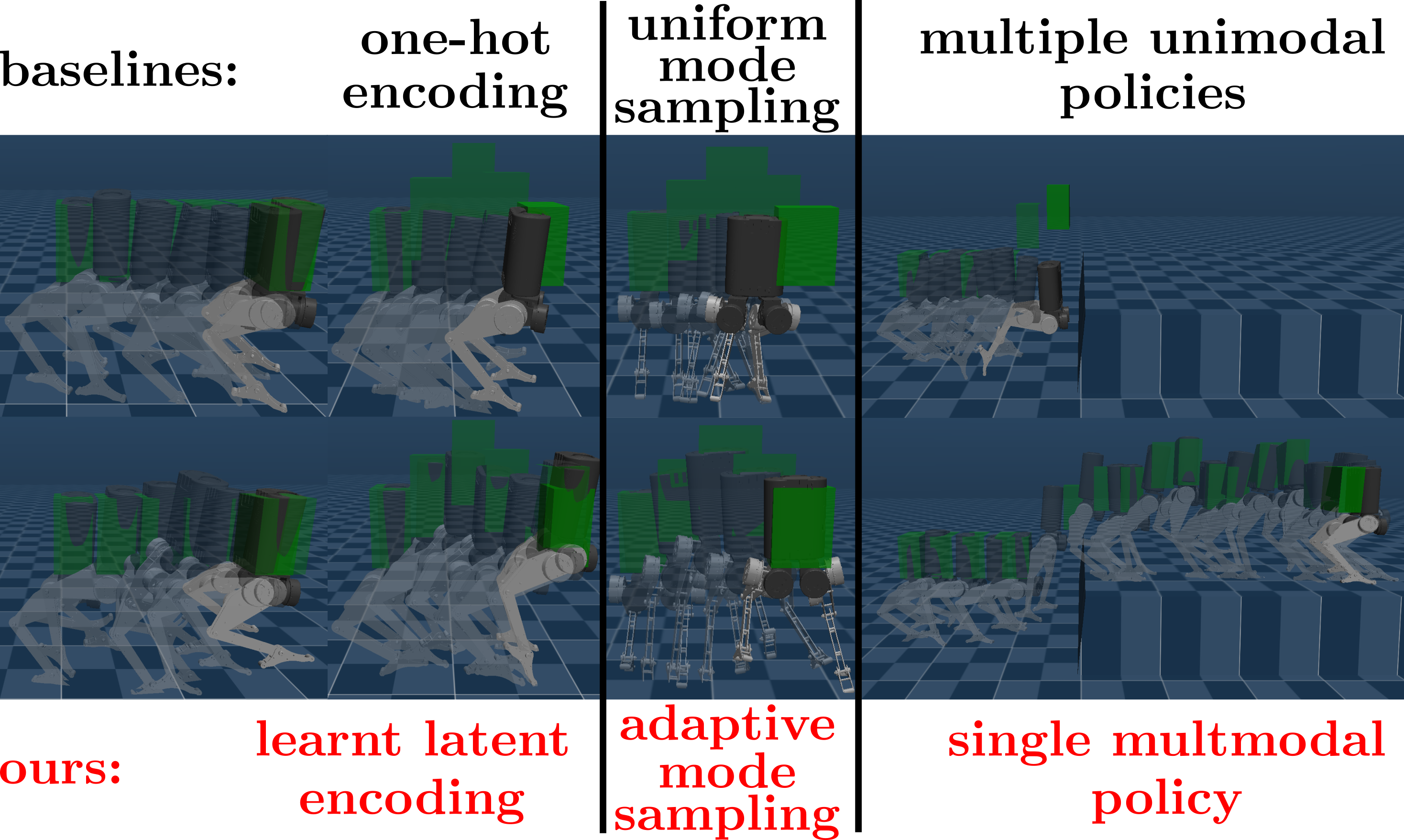}

	\caption{\textit{Simulation results of assorted comparisons between baseline (top) and our (bottom) design choices: 1) left: the distinction between walk and leap modes (mode aliasing) 2) center: failure in a subset of modes, i.e., leap-left (mode collapse) 3) right: commanded mode transitions as in Fig. \ref{fig:trans_wlaul2n} }}
	\label{fig:ma_mc_vis}
	\vspace{-5mm}
\end{figure}

\begin{figure*}[t]
	\centering

	\includegraphics[width=58mm]{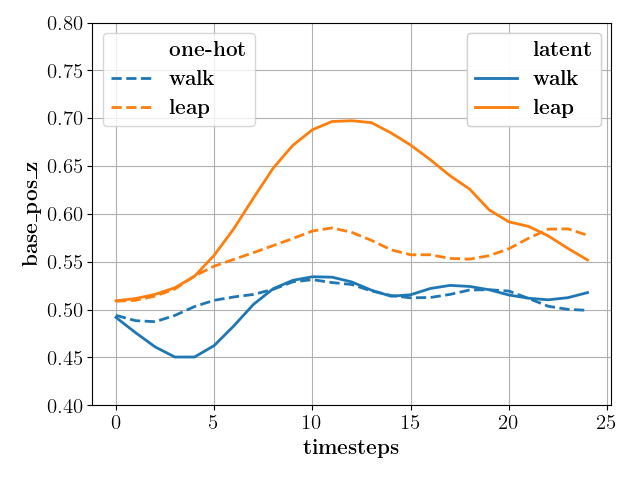}
   	\includegraphics[width=58mm]{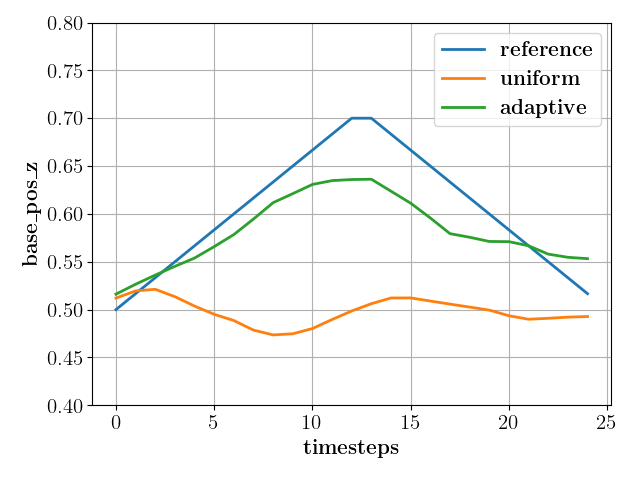}
        \includegraphics[width=58mm]{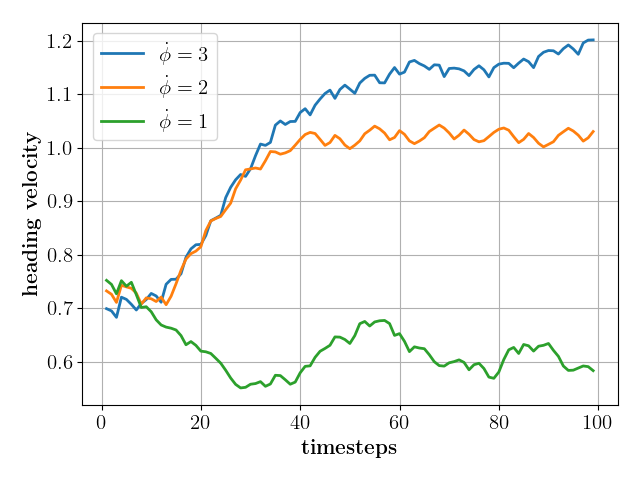}
	\caption{\textit{Figure showing mode aliasing (left), mode collapse (centre) and the effect of clock rate on heading velocity (right)}}
	\label{fig:mode_alias_colap_dphase}
	\vspace{-6mm}
\end{figure*}

\textbf{Choice of behavior encoding:} Similar to \cite{sharma19, Vollenweider2022}, for a baseline, we train by uniformly sampling the different locomotion modes in each episode that were parametrized as latent vectors with one-hot encoding. We compare our approach with the baseline by keeping everything the same and swapping the parameterization for learnt latent encodings. A key observation was that the policies learnt with one-hot encoding resulted in indistinguishable locomotion modes, i.e., mode aliasing. The attached video and Fig. \ref{fig:ma_mc_vis} (left) show that the baseline policy results in a similar base height profile for both walk and leap modes. On the other hand, policy trained with learnt latent encoding Fig. \ref{fig:mode_alias_colap_dphase} (left), results in a distinct motion for each mode where there is a well-defined rise and fall of the base height for the leap mode while it is kept constant at the nominal value for the walk mode, as desired. We hypothesize that the features in our latent encoding reflect crucial attributes of the locomotion mode that aids the policy in realizing discernible behaviors successfully.
\begin{table}
\centering
  \begin{tabular}{|c|c|c|c|}
    \hline

    \multirow{2}{*}{policy}
    & \multirow{2}{*}{sampling technique}
    & norm. MRA  & norm. MRA    \\
    
    &          & all modes & all transitions \\

\hline

    $\pi_1$ ($9$ modes, & uniform & $0.81\pm0.06$ & $0.71\pm0.09$    \\
    \cline{2-4}
    $81$ transitions)& adaptive & $0.87\pm0.04$ & $0.77\pm0.05$    \\

\hline

    $\pi_2$ ($4$ modes, & uniform & $0.87\pm0.1$ & $0.64\pm0.18$    \\
    \cline{2-4}
    $16$ transitions)& adaptive & $0.91\pm0.05$ & $0.79\pm0.24$    \\

\hline



  \end{tabular}
  \caption{Normalized Mean Returns Across (MRA) all modes and all transitions for uniform vs adaptive sampling}
\label{table:mpamt}

  \vspace{-4mm}
\end{table}

\begin{figure}
	\centering
   	\includegraphics[width=\linewidth]{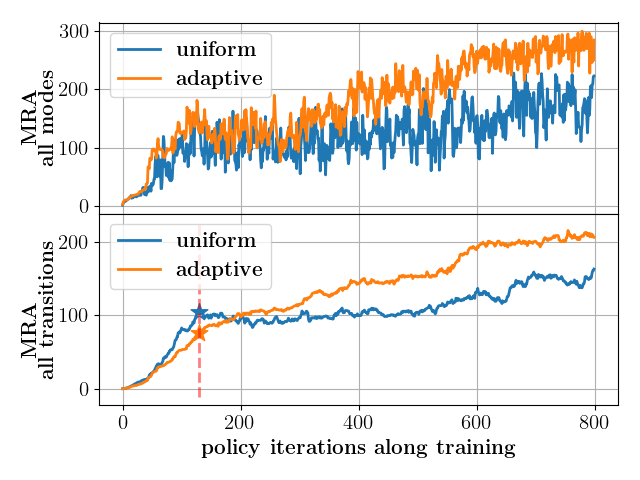}	
	\caption{\textit{Mean Returns Across (MRA) various modes (top) and transitions (bottom) along training}}
	\label{fig:mrt_mt}
	\vspace{-4mm}
\end{figure}

\textbf{Choice of mode sampling:} As mentioned, all three experiments were conducted and trained with uniform (baseline) and adaptive (proposed) sampling. Pertaining to the modes, Fig. \ref{fig:mrt_mt} (top) shows that the policies trained with adaptive mode sampling have a higher mean return across all modes as compared to sampling modes uniformly along the training. This effect is particularly significant, for a higher number of modes, as in the case shown in Fig. \ref{fig:mrt_mt}, which was of $\pi_1$ with $9$ modes in training. Furthermore, uniform sampling often leads to mode collapse, where the policy only succeeds in a subset of modes while failing or performing poorly in the rest, which can be seen in Fig. \ref{fig:ma_mc_vis}(center). $\pi_1$ trained with uniform sampling was successful in every other mode in the training but leap-left mode where the resulting base height profile fails to be anywhere close to the reference motion shown in blue in Fig. \ref{fig:mode_alias_colap_dphase} (center). In contrast, the adaptive sampling variant successfully imitates the reference to the best extent permitted by the robot's dynamics across all modes, as seen in the attached video and Fig. \ref{fig:mode_alias_colap_dphase} (center).

\textbf{Trained with and without contact sequence:} From $\pi_1$, $\pi_2$, we can observe that a rough demonstration with just the keyframes for the base pose is sufficient to realize the locomotion modes without needing a reference of either the joint or contact states. The attached video shows that this training leads to asymmetric yet natural-looking gaits for periodic modes like walk and leap. However, we often need well-defined contact sequences to locomote on constrained terrains such as stepping stones. To this end, we illustrate the flexibility of our framework by integrating a rough demonstration of the contact state along with the base pose, in $\pi_3$. Particularly, with the same base pose reference of constant base height, we construct two modes: hop and walk, by only changing the reference contact states between the legs to be in and out of synchronization, respectively. The video results of $\pi_3$ show the resultant symmetric well-defined walking and hopping behavior upon using a contact sequence reference. 

\textbf{Frequency of motion in modes:} The current framework with mode-conditioned policies lacks granular control over certain states ( e.g., walking velocity), unlike the current SOTA command-conditioned policies and MPCs. Though we justify this trade-off by realizing fruitful high-level tasks through mode planning, preliminary investigations show that we can crudely control the frequency or rate of motion by modulating the clock rate of the policy's clock input. Specifically, in the leap mode, increasing the clock signal frequency led to a direct increase in the heading speed of the robot. In Fig. \ref{fig:mode_alias_colap_dphase} (right), the heading velocity is double ($1.0$ m/s) the nominal value of $0.5$ m/s when we increase the clock rate by a factor of $2$. This effect saturates at thrice the nominal clock rate at a velocity of $\sim1.2$ m/s, which we believe is due to the physical limits of the robot's form factor and the nature of the given leap mode.






\subsection{Performance across all the inherent transitions}

As stated before, the major advantage of learning a single policy to operate on multiple locomotion modes lies in inherently realizing smooth inter-mode transition maneuvers. We elucidate the above hypothesis and the benefit of training with adaptive transition sampling below


\begin{figure}
   	\includegraphics[width=\linewidth]{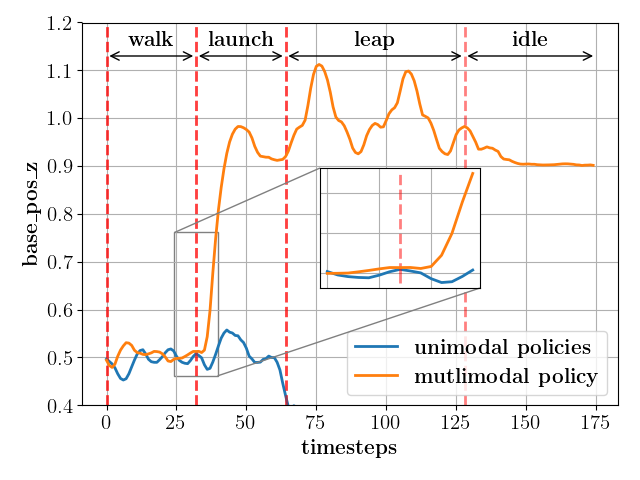}	
	\caption{\textit{Comparison of base height profiles for motion shown in Fig. \ref{fig:ma_mc_vis} (right), with mode  transitions: walk $\rightarrow$ launch, launch $\rightarrow$ leap, and leap $\rightarrow$ idle }} 
	\label{fig:trans_wlaul2n}
	\vspace{-7mm}
\end{figure}
\textbf{Emergent non-trivial transitions:} We first showcase the significance of learning a single multimodal policy with implicit transitions as opposed to having multiple unimodal policies and explicit transitions. For the baseline akin to current SOTA, we train $4$ different unimodal policies that specialize in a single mode, namely: walk, leap, launch, and idle, respectively, and verified that these policies successfully realized their respective modes when tested without any transitions. However, when we try to compose motions with transitions such as pacing and jumping on top of a box, the optimal transition time and maneuver are not straightforward to obtain. Hence, when we discreetly switch between the walking and launch unimodal policies, the baseline fails as the states faced during the transition are out of distribution for the launch policy, as seen in the attached video and Fig. \ref{fig:ma_mc_vis} (right). In contrast,  Fig. \ref{fig:trans_wlaul2n} shows that the multimodal policy results in a feasible and emergent transition maneuver (zoomed-in view) to perform the parkour-like task with the transitions: walk $\rightarrow$ launch, launch $\rightarrow$ leap, and leap $\rightarrow$ idle. Thus, the multimodal policy successfully stitches together a smooth sequence of distinct locomotion modes even with discrete switching commands.

\textbf{Choice of transitions sampling:} Similar to the case of mode sampling, uniformly sampling transitions in training leads to transition collapse, thus failing in a subset of all possible transitions. To circumvent this, extending the adaptive sampling to transitions, we see higher mean returns over all the transitions than uniform sampling along training in Fig. \ref{fig:mrt_mt} (bottom). Furthermore, we can also see that uniform sampling falls into a local optimum where it is successful in specific transitions but sub-optimal in the rest leading to high returns early on in training (point denoted with a star in Fig. \ref{fig:mrt_mt} (bottom) ). However, it never recovers from this local optima and thus results in transition collapse. In contrast, adaptive sampling of transitions ensures consistent performance across all transitions and increases along the training, leading to compelling and reactive transition maneuvers, as seen in the supplementary video and table \ref{table:mpamt}.


\subsection{ Rapid mode-planning for high-level tasks}

\begin{figure}
	\centering
   	\includegraphics[width=\linewidth,height=5.5cm]{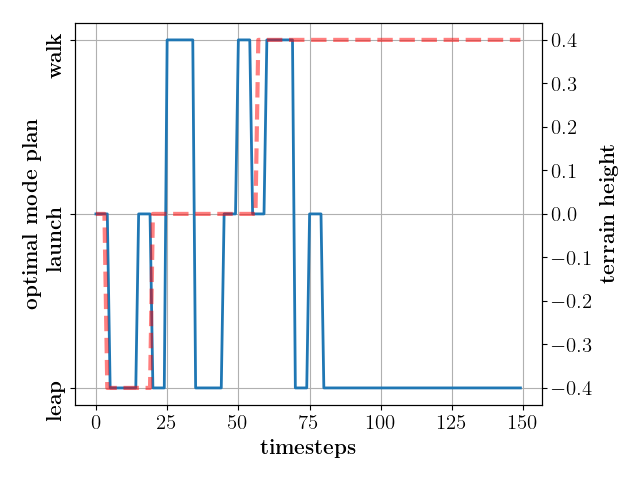}	
	\caption{\textit{An optimal open-loop mode plan (blue) to cross a track with a gap and a block with the depicted height (red) }}
	\label{fig:mode_plan}
\end{figure}

\begin{table}
  \centering
  \begin{tabular}{|c|c|c|c|c|}
    \hline

    \multirow{2}{*}{Experiment} & $p_x^{goal}$ & \multirow{2}{*}{$k$} & $dt$ & solve time  \\
    & (m) & &(sec)& (mins:sec) \\

\hline
     reach $p_x^{goal}$ with a & \multirow{2}{*}{[2.0, 0.0, 0.5]} & \multirow{2}{*}{11} & \multirow{2}{*}{0.3}& \multirow{2}{*}{3:05$\pm$0:08}  \\ 
     gap in between  &   & & &\\
\hline
     reach $p_x^{goal}$ with a & \multirow{2}{*}{[2.0, 0.0, 0.5]} & \multirow{2}{*}{11} & \multirow{2}{*}{0.3}& \multirow{2}{*}{3:33 $\pm$ 0:10}  \\ 
     plateau in between  &   & & &\\
\hline
     reach $p_x^{goal}$ with a gap & \multirow{2}{*}{[2.0, 0.0, 0.9]} & \multirow{2}{*}{30} & \multirow{2}{*}{0.15}& \multirow{2}{*}{2:41 $\pm$ 0:20}  \\ 
    and block in between  &   & & &\\
       
\hline

  \end{tabular}
  \caption{Summary of experiments for mode planning}
\label{table:qplans}

  \vspace{-5mm}
\end{table}
As presented in the attached video, we can successfully traverse challenging terrains (as in Fig. \ref{fig:diff_mode_motion_trace}) with discreetly-switched open-loop plans (as in Fig. \ref{fig:mode_plan}) generated by the mode planner. The planner generates agile parkour-like motions by exploiting the multimodal policy's different locomotion modes to overcome gaps of width $0.45$ m, plateaus of height $0.2$ m, and blocks of height $0.4$.
The experiment shown in Fig. \ref{fig:diff_mode_motion_trace} demonstrates a leap of width $102 \%$ the maximum leg length and a jump of height $80 \%$ the nominal COM height of our mini-biped platform. 
As seen in table \ref{table:qplans}, thanks to the low-dimensional and compact Q-table, we can rapidly compose these complex behavior compositions  (as in Fig. \ref{fig:diff_mode_motion_trace} with the corresponding plan in Fig. \ref{fig:mode_plan})  in a few minutes, thus being capable of planning in real-time.

\section{Conclusion}
\label{sec:conc}

This paper presents a novel framework for learning multimodal bipedal locomotion with periodic, steady-state, and transient behaviors from rough reference motion clips. A single policy trained with learnt latent modes and adaptive sampling has been shown to consistently perform diverse locomotion skills and successful transitions despite their varied complexities by alleviating mode collapse and aliasing. Effectively caching solutions across distinct regions of the state space, the multimodal policy guarantees feasible transition maneuvers, thereby radically decreasing the compute load and solving time of a high-level planner to compose smooth multimodal behaviors. Simulation results show that a primitive open-loop task-based planner, in conjugation with the trained multimodal policy, can result in agile parkour-like behaviors to traverse challenging terrains with gaps, plateaus, and blocks. Since transferability to hardware was not a focus of this work, extensive reward shaping to improve the quality of the motion (to avoid high impact, jittery motions, etc) was absent, which is a work in progress. Future work will include developing a sophisticated mode planner utilizing exteroceptive feedback of the environment for tackling contact-rich loco-manipulation tasks. 


 \bibliographystyle{ieeetr}
\bibliography{references}
\end{document}